\documentclass[twocolumn, conference]{IEEEtran}
\IEEEoverridecommandlockouts
\usepackage{cite}
\usepackage{amsmath,amssymb,amsfonts}
\usepackage{amsthm}
\usepackage{mathtools}
\usepackage{algorithmic}
\usepackage{graphicx}
\usepackage{textcomp}
\usepackage{xcolor}
\usepackage{tabularx}
\usepackage{wrapfig}
\usepackage{enumitem}
\usepackage{mathtools}
\usepackage{multirow}
\usepackage{float}
\usepackage{algorithm,algorithmic}
\theoremstyle{plain}

\theoremstyle{definition}

\newcolumntype{L}[1]{>{\raggedright\arraybackslash}p{#1}}
\newcolumntype{C}[1]{>{\centering\arraybackslash}p{#1}}

\global\long\def\ER{Erd\"{o}s-R\'{e}nyi}%

\def\BibTeX{{\rm B\kern-.05em{\sc i\kern-.025em b}\kern-.08em
    T\kern-.1667em\lower.7ex\hbox{E}\kern-.125emX}}

\providecommand{\definitionname}{Definition}
\providecommand{\theoremname}{Theorem}

\begin{document}

\title{Graph Coding for Model Selection and Anomaly Detection in Gaussian Graphical Models
}

\author{
\IEEEauthorblockN{Mojtaba Abolfazli, Anders H{\o}st-Madsen, June Zhang\thanks{Supported in part by NSF grant CCF-1908957.}}
\IEEEauthorblockA{Department of Electrical Engineering \\
University of Hawaii, Honolulu, HI, USA \\
Email: \{mojtaba, ahm, zjz\}@hawaii.edu}
\and\IEEEauthorblockN{Andras Bratincsak}
\IEEEauthorblockA{Department of Pediatrics, John A. Burns School of Medicine\\
University of Hawaii, Honolulu, HI, USA \\
Email: andrasb@hphmg.org}
}

\maketitle
\setlength{\abovedisplayskip}{5pt}
\setlength{\belowdisplayskip}{5pt}
\setlength{\textfloatsep}{5pt}

\begin{abstract}
A classic application of description length is for model selection with the minimum description length (MDL) principle. The focus of this paper is to extend description length for data analysis beyond simple model selection and sequences of scalars. More specifically, we extend the description length for data analysis in Gaussian graphical models. These are powerful tools to model interactions among variables in a sequence of i.i.d Gaussian data in the form of a graph. Our method uses universal graph coding methods to accurately account for model complexity, and therefore provide a more rigorous approach for graph model selection. The developed method is tested with synthetic and electrocardiogram (ECG) data to find the graph model and anomaly in Gaussian graphical models. The experiments show that our method gives better performance compared to commonly used methods.

\end{abstract}

\section{Introduction}
Multivariate Gaussian distributions are widely used in modeling real-world problems where the relationship between approximately normally distributed variables is of interest. Examples are the distribution of stock returns \cite{kon1984models, golosnoy2012conditional}, protein-protein interactions \cite{baldassi2014fast}, and brain functional activities \cite{varoquaux2010brain}.

Let $X = (X_1, \ldots, X_p)$ be a $p-$dimensional random vector with multivariate Gaussian distribution $\mathcal{N}(\boldsymbol{0},\,\boldsymbol{\Sigma})$. The inverse of covariance matrix, $\boldsymbol{\Sigma}^{-1} = \boldsymbol{\Omega}$, is known as the precision matrix. If the $(i,j)$ entry of the precision matrix is $0$, then $X_i$ and $X_j$ are conditionally independent given all the other variables. Therefore, the conditional independence relationships of a multivariate Gaussian can be visualized by an unweighted, undirected graph, $G=(V,E)$ whose adjacency matrix, $\boldsymbol{A}$ is such that 

\begin{equation}
\label{eq:Adj}
\mathbf{A}_{ij}=
\begin{cases}
1 \hspace{0.5cm} &\text{if} \hspace{0.3cm} \boldsymbol{\Omega}_{ij}\neq 0, i \neq j \\
0 \hspace{0.5cm} &\text{otherwise}.
\end{cases}
\end{equation}

\begin{figure}[t]
	\begin{centering}
  \includegraphics[width=3.1 in]{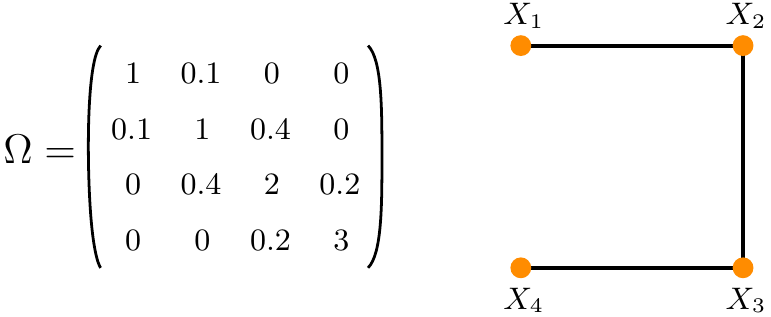}
  \caption{\label{GGM.fig}Illustration of conditional independence graph $G$ associated with a precision matrix.}
 \end{centering}
\end{figure}

The graph $G$ is known as the conditional independence graph. The sparsity pattern of $\boldsymbol{\Omega}$ determines the topology of $G$.
Figure~\ref{GGM.fig} shows an example precision matrix $\boldsymbol{\Omega}$ and its associated conditional independence graph.

Often, we are interested in the problem of estimating $G$ from i.i.d observations $\{\mathbf{x}_1, \ldots, \mathbf{x}_N\}$. The problem is especially difficult when the Gaussian distribution has high dimensionality but with few observations ($N < p$). In most approaches, the estimator of $\boldsymbol{\Sigma}$ (or $\boldsymbol{\Omega}$) contains a regularization term to 1) prevent overfitting, 2) control the sparsity of the estimated solution. In the case of estimating the precision matrix, the weight of the regularization term (as determined for example by a regularization parameter $\lambda$) has a high impact on the structure of the conditional independence graph, $G$.

In this paper we develop a new methodology for choosing the
weighting of the regularization term, i.e, $\lambda$.
Our method is based on Rissanens's minimum description length (MDL).
Rissanen's MDL principle codifies model selection, which balances between how well a model describes the given data and the complexity of the model \cite{Rissanen86}. In practice, however, it is often difficult to compute an exact description length for some statistical models. For example, previous work in model selection of Gaussian graphical models does not account for the exact topology of the conditional independence graph.

Previously, we developed methods for lossless coding of arbitrary unweighted, undirected graph structures \cite{host2018coding}. In this paper, we use the description length, the minimum number of bits required to describe the data losslessly, associated with these coding methods in conjunction with MDL principle to learn the best graph model of data. More specifically, we use the description length for model selection and anomaly detection in Gaussian graphical models.
We leverage existing work in universal graph coding \cite{ChoiSzpankowski12,host2018coding}, based on graph statistics such as edge probability, degree distribution and triangle distribution, to compute the exact description length of a Gaussian graphical model. 

An overview of previous studies for model selection and anomaly detection in Gaussian graphical models is given in section~\ref{sec:PrevStud}.
In section~\ref{sec:ModelSelec}, we use our method in model selection of graphical lasso \cite{friedman2008sparse}, a popular method for estimating a sparse precision matrix. We show that our method outperforms competing model selection methods especially in the case where the number of observations is less than the number of variables. In section~\ref{sec:Anomaly}, we use description length for anomaly detection using the principle of atypicality \cite{host2019data}. We show that atypicality can capture anomalous data by describing it with fewer bits in itself rather than the optimum coder for typical data. 
The experiment results on a real-world electrocardiogram (ECG) dataset show that our approach can find anomaly in multivariate normally distributed data with high accuracy.

\section{Previous Work} \label{sec:PrevStud}
Different estimators have been introduced to fit a sparse graphical model to multivariate Gaussian data. Reference \cite{meinshausen2006high} proposed a method that works based on neighborhood selection, where conditional independence for each node is estimated by regressing on other variables with lasso. Another popular estimator is graphical lasso \cite{friedman2008sparse} that work based on $l_1-$penalized log-likelihood. This method is easy to implement and fast to compute.
Other penalized methods have been introduced to solve the problem efficiently \cite{banerjee2008model, yuan2007model, lam2009sparsistency,  witten2011new, hsieh2014quic}.
All of these estimators have a regularization parameter $\lambda$ that controls the sparsity level of conditional independence graph ranging from fully disconnected nodes to a fully connected graph. Therefore, their performance is highly influenced by the value of regularization parameter $\lambda$.

To pick the best value of $\lambda$, different model selection techniques have been used in the literature. We can divide them into two main classes: information-based methods and resampling methods. Information-based methods such as Bayesian information criteria (BIC)  \cite{yuan2007model}, Akaike information criteria (AIC) \cite{menendez2010gene}, and extended Bayesian information criteria (EBIC) \cite{foygel2010extended} work based on the performance on training dataset and the complexity of the model.
This class offers a good statistical interpretation by adding a penalty term to compensate for the overfitting problem of more complex models. 
EBIC uses an extra penalization parameter that controls preference to simpler models and is set manually based on the cautious and application.

Resampling methods work based on measuring the performance on out-of-sample data. They split the data into a subset of samples to fit a model and the remaining samples to estimate the efficacy of the model. This process is repeated multiple times, and the results are aggregated to find the best model. The most common method in this class is cross-validation (CV) \cite{rothman2008sparse}. Other methods are Generalized Approximate Cross-validation (GACV) \cite{lian2011shrinkage}, Rotation Information Criterion (RIC) \cite{lysen2009permuted}, and Stability Approach to Regularization Selection (StARS) \cite{liu2010stability}.
The major shortcoming of resampling methods is their high computational cost since they require to solve the problem across all sets of subsamples.

Besides these two approaches, there are several studies that propose other methods. 
The authors in \cite{vujavcic2015computationally} suggested a method based on KL loss that improves the performance of BIC when the sample size
is small. Also, several methods based on graph structure were proposed in \cite{mestres2018selection} considering there is a high-level knowledge about graph structure. The paper \cite{wang2010information} provides information-theoretic bounds for correct recovery of conditional independence graph based on sample size, number of variables, and the maximum node degree.

All of the aforementioned methods are missing a key aspect of graphical models. They do not consider the graph structure as an informative part of model selection. For example, EBIC only considers the number of edges in estimated graph structure while one may argue that we can have different graph structures with the same number of edges but various degrees of complexity. In this paper, we want to consider the whole graph structure for data analysis in Gaussian graphical models.

In addition to using description length for model selection, we will show another application for anomaly detection in Gaussian graphical models. Previous studies for anomaly detection in Gaussian graphical models are limited to changepoints detection in graph structures over time. 
A time-varying graphical lasso method was developed in \cite{hallac2017network} by imposing a temporal penalty term that limits the evolution of network structure over time. Reference \cite{avanesov2018change} introduced a multiscale method for changepoint detection that attains a trade-off between sensitivity and accuracy. In \cite{maurya2018contrastive}, structural changes are determined based on the difference between learned background precision matrix and sliding foreground precision matrix.
Our approach is different as it uses description length for anomaly detection in data; that is, we encode data using a lossless source coder, and use the resulting codelength as the decision criteria.

\section{Model Selection} \label{sec:ModelSelec}
In this section, we apply graph-coding techniques introduced in \cite{host2018coding, ChoiSzpankowski12} for model selection in Gaussian graphical models. In other words, our goal is to identify zeros in the precision matrix.
For that, we use MDL principle and select $\lambda$ that minimizes the sum of description length of the graph structure and data under that graph structure
\begin{equation}
    \arg \min_{\lambda} L(G_{\lambda}) + L(D|G_{\lambda})
    \label{eq:MDL}
\end{equation}
where $L(G_{\lambda})$ is the description length of conditional independence graph structure $G_{\lambda}$ that depends on $\lambda$, and $L(D|G_{\lambda})$ denote the description length of data $D$ when encoded with $G_{\lambda}$. Our results show that $G_{\lambda}^*$ associated with $\lambda^*$ minimizing \eqref{eq:MDL} gives a better model of data compared to other methods such as CV, BIC, and EBIC. 

In the following, we provide the details on how to compute terms in \eqref{eq:MDL}.

\subsection{Encoding Graph Structure \label{sec:GraphCoding}}
To compute $L(G_{\lambda})$, we first need to find the underlying conditional independence graph $G_{\lambda}$. By applying an estimator to the sequence of $N$ i.i.d observations $\{\mathbf{x}_1, \ldots, \mathbf{x}_N\}$ from $p-$variate Gaussian distribution, we obtain an estimated precision matrix $\boldsymbol{\hat{\Omega}}$ for each realization of $\lambda$. We have dropped the explicit 
dependence on $\lambda$ from $\boldsymbol{\hat{\Omega}}$ and other $\lambda-$related parameters for the sake of simplicity of notation.

In this paper, we use graphical lasso \cite{friedman2008sparse}, which is one of the most commonly used estimators in Gaussian graphical models. Our approach can be easily applied to any other estimator.
The estimated precision matrix $\boldsymbol{\hat{\Omega}}$ obtained from the graphical lasso estimator is used to form the adjacency matrix $\mathbf{A}$ of the unweighted, undirected graph $G_{\lambda}$ as described in \eqref{eq:Adj}.
Next, we can use any of the following graph coders to compute $L(G_{\lambda})$:
\begin{itemize}
    \item \textbf{Degree distribution:} This coder uses the degree distribution of an underlying graph for encoding \cite{host2018coding}.
    \item \textbf{Triangle:} This coder looks for triangles to encode a graph structure \cite{host2018coding}.
    \item \textbf{IID structure:} This coder utilizes the edge probability to encode a graph structure. It is designed for \ER\ graphs and uses a two-stage compression algorithm  \cite{ChoiSzpankowski12}. First, the structure of graph is converted into two binary strings. Then, an arithmetic encoder is used to compress these strings. The decoder can reconstruct the graph that is isomorphic to the original graph.
\end{itemize}

For more details about these coders, an interested reader can see \cite{host2018coding, ChoiSzpankowski12}.

\subsection{Encoding Data with Graph Structure \label{sec:DadtaCoding}}
Computing $L(D|G_{\lambda})$ is not as straightforward as computing $L(G_{\lambda})$. There are two challenges in this regard. First, we need to deal with real-valued data $\{\mathbf{x}_1, \ldots, \mathbf{x}_N\}$ where lossless source coding is not generalized directly. Second, we have to encode the data based on the underlying conditional independence graph $G_{\lambda}$. To encode real-valued data, we assume a fixed-point representation with a (large) finite number, $r$, bits after the period, and an unlimited number of bits before the period \cite{Rissanen83}. Therefore, the number of bits required to represent data $x$ according to the pdf distribution $f(x)$ is given by
\begin{align}
L(x) &= - \log \int _{x}^{x+2^{-r}} f(t) dt \approx - \log \left(f(x)2^{-r} \right) \nonumber \\
 &= - \log f(x) + r
\end{align}
Since we are only interested in relative codelengths between different models, the dependency on $r$ cancels out.

The second challenge is how to encode the data with respect to the graph structure. As mentioned, the data is generated by multivariate Gaussian distribution, which can be characterized by covariance matrix $\boldsymbol{\Sigma}$. Since we do not have access to true covariance matrix $\boldsymbol{\Sigma}$, we can find an estimate $\boldsymbol{\check{\Sigma}}$ based on the sample covariance matrix $\boldsymbol{S}$. It should be noted that the structure of $\boldsymbol{\check{\Sigma}}^{-1}$ should match with the structure of $G_{\lambda}$ obtained from the previous step:
\begin{equation} \label{demp.eq}
\begin{cases}
\boldsymbol{\check{\Sigma}}_{ij} = \boldsymbol{S}_{ij} &\text{if} \hspace{0.3cm} i=j \ \text{or} \ \boldsymbol{A}_{ij} \neq 0 \\
\boldsymbol{\check{\Sigma}}_{ij}^{-1} = 0 \hspace{0.5cm} &\text{otherwise}.
\end{cases}
\end{equation}
This problem is known as \emph{matrix completion} problem and is widely studied in the literature \cite{grone1984positive}. Dempster in \cite{dempster1972covariance}, proved the existence and uniqueness of maximum likelihood solution for $\boldsymbol{\check{\Sigma}}$ when the sample covariance matrix $\boldsymbol{S}$ is positive definite. 
He presented a coordinate-descent algorithm to find the solution iteratively.
It should be noted that $\boldsymbol{\hat{\Omega}}$ is the precision matrix obtained from the graphical lasso solution and is not necessarily the inverse of estimated covariance matrix $\boldsymbol{\check{\Sigma}}$.

Once we estimate $\boldsymbol{\check{\Sigma}}$, we use predictive minimum description length (MDL) \cite{Rissanen86} to compute the codelength of data under the graph structure:
\begin{equation}
    L(D | G_{\lambda}) = -\sum_{i=1}^{N-1} \log f\left(\mathbf{x}_{i+1}|\boldsymbol{\hat{\theta}}(\mathbf{x}_1,\ldots, \mathbf{x}_i)\right) \label{eq:predMDL}
\end{equation}
where $f(\cdot|\cdot)$ is the conditional probability and $\boldsymbol{\hat{\theta}}(\mathbf{x}_1,\ldots, \mathbf{x}_i)$ denote the maximum likelihood estimate of the parameter, which in this case is the estimated covariance $\boldsymbol{\check{\Sigma}}$ obtained under $G_{\lambda}$. The codelength in \eqref{eq:predMDL} is a valid codelength since it is sequentially decodable. It does not work for the first few samples, as there is no estimate of the covariance matrix. Instead, we encode the first few samples with a default distribution that is the same among different realizations of $\lambda$. 

Finally, the best conditional independence graph structure $G_{\lambda}$ is obtained by minimizing the summation of codelength of $G_{\lambda}$ and data encoded under $G_{\lambda}$. In the following, we present an algorithm to find the best graph model of data $G_{\lambda^*}$ associated with $\lambda^*$

\begin{algorithm}
	\caption{Find the best graph model of data $G_{\lambda^*}$ via graph-coding methods}
	\label{Alg}
	\hspace*{\algorithmicindent} \textbf{Input: }Samples $\{\mathbf{x}_1, \ldots, \mathbf{x}_N\} \sim \mathcal{N}(\boldsymbol{0},\,\boldsymbol{\Sigma})$ and set of candidate regularization parameters $\{\lambda_1,\ldots, \lambda_K\}$. \\
    \hspace*{\algorithmicindent} \textbf{Output:} Best graph model of data $G_{\lambda^*}$.
    
    \begin{algorithmic}[1]
		\FOR {Each realization of $\lambda \in \{\lambda_1,\ldots, \lambda_K\}$.}
		\STATE {Apply graphical model estimator to find $\boldsymbol{\hat{\Omega}}$ and build $G_{\lambda}$ based on \eqref{eq:Adj}.}
		\STATE {Use any graph coder in Section~\ref{sec:GraphCoding} to compute $L(G_{\lambda})$.} 
		\STATE {Compute $L(D|G_{\lambda})$ by using \eqref{eq:predMDL} where $\boldsymbol{\hat{\theta}}(\mathbf{x}_1,\ldots, \mathbf{x}_i)$ denote $\boldsymbol{\check{\Sigma}}$ obtained from \eqref{demp.eq}.}
		\STATE {Add up codelengths resulted from step~(3) and step~(4).}
		\ENDFOR
		\RETURN {$G_{\lambda^*}$ with shortest total codelength in step~(5).}
	\end{algorithmic}
\end{algorithm}

\subsection{Experiments}\label{sec:ModelExp}
We now provide simulation results on synthetic data generated from zero-mean multivariate Gaussian distribution with known precision matrix, $\boldsymbol{\Omega}$. We will compare the performance of our approach against other methods in the recovery of conditional independence $G$. We use the F1-score as it is a widely used metric. F1-score is the harmonic mean of precision and recall where the precision is the ratio of the number of correctly estimated edges to the total number of edges in the estimated graph, and the recall is the ratio of the number of correctly estimated edges to the total number of edges in the true graph \cite{liu2010stability}.
We consider two cases: 1) $N>p$ with $N/p = 2$ and 2) $N<p$ with $N/p = 0.5$. The experiments were repeated for $p = 100, 200$.
We applied the graphical lasso method described in \cite{friedman2008sparse} for 50 realizations of $\lambda$ in the range  $[\bar{\lambda}/10, \bar{\lambda}]$ where $\bar{\lambda}$ is the minimum value of $\lambda$ that gives totally isolated nodes. 
The multivariate Gaussian data was generated with different precision matrix structures have been frequently considered as test cases in the literature \cite{li2010inexact, tan2014learning}:
\begin{itemize}
    \item Cycle structure with $\boldsymbol{\Omega}_{ii} = 1,\ \boldsymbol{\Omega}_{i,i-1} = \boldsymbol{\Omega}_{i-1,i} = 0.5,\ \boldsymbol{\Omega}_{1p} = \boldsymbol{\Omega}_{p1} = 0.4$. 
    
    \item Autoregressive process of order one AR(1) with $\boldsymbol{\Omega}_{ii} = 1,\ \boldsymbol{\Omega}_{i,i-1} = \boldsymbol{\Omega}_{i-1,i} = 0.5$.
    
    \item Autoregressive process of order two AR(2) with $\boldsymbol{\Omega}_{ii} = 1,\ \boldsymbol{\Omega}_{i,i-1} = \boldsymbol{\Omega}_{i-1,i} = 0.5, \ \boldsymbol{\Omega}_{i,i-2} = \boldsymbol{\Omega}_{i-2,i} = 0.25$.
    
    \item \ER\ (ER) structure with $\boldsymbol{\Omega}_2 = \boldsymbol{\Omega}_1 + \delta \boldsymbol{I}_p$ where $\boldsymbol{\Omega}_1$ is a matrix with off-diagonal elements taking values randomly chosen from uniform distribution $\mathcal{U}~(0.4,0.8)$ with the probability of $2/p$ and diagonal values set to zero. To keep $\boldsymbol{\Omega}_2$ positive definite, we choose $\delta = \rho +0.05$ where $\rho$ is the absolute value of the minimum eigenvalue of $\boldsymbol{\Omega}_1$. Here $\boldsymbol{I}_p$ is the identity matrix of size $p$.
    
    \item Hub structure with two hubs. First, we create the adjacency matrix  $\boldsymbol{A}$ by setting off-diagonal elements to one with the probability of $0.01$ and zero otherwise. Next, we randomly select two hub nodes and set the elements of the corresponding rows and columns to 1 with the probability of $0.7$ and zero otherwise. After that for each nonzero element $\boldsymbol{A}_{ij}$, we set $\boldsymbol{A}_{ij}$ with a value chosen randomly from uniform distribution $\mathcal{U}~(-0.75,-0.25) \cup (0.25,0.75) $. Then, we set $\boldsymbol{\Omega}_1 = \frac{1}{2}(\boldsymbol{A}+\boldsymbol{A}^T)$. The final precision matrix $\boldsymbol{\Omega}_2$ is obtained by $\boldsymbol{\Omega}_2 = \boldsymbol{\Omega}_1 + \delta \boldsymbol{I}_p$ as set-up for \ER\ graph.
\end{itemize}

Table~\ref{LowDim.tab} and Table~\ref{HighDim.tab} show the results of applying different methods to recover the conditional independence graph by applying  graphical lasso as the estimator for $N > p$ and $N<p$ cases, respectively. The results for CV are given for 5-fold CV, and for EBIC method are given for recommended value of $\gamma = 0.5$ \cite{foygel2010extended}. The results for graph-coding methods are obtained by following the steps outlined in Algorithm~\ref{Alg}.
The values are the average of 50 Monte Carlo trials and the best value (values) at each row is (are) boldfaced.
It can be seen that graph-coding techniques (columns named with Degree, IID, and Triangle) outperform other methods particularly when $N<p$.
We observed that coding with degree distribution and triangle give better performance than coding with IID structure in most of the cases. This finding confirms the necessity of having graph coders with the capability to reflect more information about the graph in their codelength.

\begin{table}[tbh]
\begin{center}
\setlength{\tabcolsep}{4pt}
\renewcommand{\arraystretch}{1.4}
\begin{tabular}{ |c|c|c|c|c|c|c|c|c| } 
 \hline
 \multirow{2}{*}{Type} & \multirow{2}{*}{$p$} & \multirow{2}{*}{$N$} &\multicolumn{3}{c|}{Benchmark methods} & \multicolumn{3}{c|}{Graph-coding methods}  \\
 \cline{4-9}
 &  & & CV & BIC & EBIC & Degree & IID & Triangle \\
 \hline
 Cycle & 100 & 200 & 0.26 & 0.65 & 0.75  & \textbf{1} & \textbf{1} & \textbf{1} \\ 
 Cycle & 200 & 400 & 0.24 & 0.62 & 0.69  & \textbf{0.99} & \textbf{0.99} & \textbf{0.99} \\ 
 
 AR(1) & 100 & 200 & 0.25 & 0.64 & 0.76  & \textbf{0.99} & 0.88 & 0.98 \\
 AR(1) & 200 & 400 & 0.24 & 0.62 & 0.70  & \textbf{1} & 0.99 & 0.99 \\ 
 
 AR(2) & 100 & 200 & 0.28 & \textbf{0.64} & na    & \textbf{0.64} & 0.62 & 0.63 \\ 
 AR(2) & 200 & 400 & 0.24 & 0.71 & na     & 0.71 & 0.68 & \textbf{0.72} \\ 
 
 ER & 100 & 200 & 0.28 & 0.63 & 0.51   & \textbf{0.67} & 0.60 & \textbf{0.67} \\ 
 ER & 200 & 400 & 0.27 & 0.68 & \textbf{0.76}   & \textbf{0.76} & \textbf{0.76} & \textbf{0.76} \\ 
 
 Hub & 100 & 200 & 0.23 & 0.47 & 0.01    & \textbf{0.49} & 0.47 & 0.48 \\ 
 Hub & 200 & 400 & 0.24 & \textbf{0.44} & 0.20     & \textbf{0.44} & 0.42 & \textbf{0.44} \\ 
 \hline
\end{tabular}
\end{center}
\caption{\label{LowDim.tab}F1-score of conditional independence graph recovery using different model selection methods.  Results are the average over 50 replications when $N > p$.}
\end{table}

\begin{table}[tbh]
\begin{center}
\setlength{\tabcolsep}{4pt}
\renewcommand{\arraystretch}{1.4}
\begin{tabular}{ |c|c|c|c|c|c|c|c|c| } 
 \hline
 \multirow{2}{*}{Type} & \multirow{2}{*}{$p$} & \multirow{2}{*}{$N$} &\multicolumn{3}{c|}{Benchmark methods} & \multicolumn{3}{c|}{Graph-coding methods}  \\
 \cline{4-9}
 &  & & CV & BIC & EBIC & Degree & IID & Triangle \\
 \hline
 
 Cycle & 100 &  50 & 0.27  & 0.26 & 0.75    & \textbf{0.89} & 0.79 & \textbf{0.89}\\ 
 Cycle & 200 &  100 & 0.23 & 0.29 & 0.68  & \textbf{0.96} & 0.80 & \textbf{0.96} \\ 
 
 AR(1) & 100 & 50 & 0.28  & 0.26 & 0.73    & \textbf{0.89} & 0.80 & 0.88 \\ 
 AR(1) & 200 & 100 & 0.23 & 0.29 & 0.70  & \textbf{0.95} & 0.78 & 0.94 \\ 
 
 AR(2) & 100 & 50 & 0.32  & 0.30 & na     & 0.36 & \textbf{0.37} & 0.36 \\ 
 AR(2) & 200 & 100 & 0.27 & 0.33 & na     & 0.52 & \textbf{0.53} & 0.52 \\ 
 
 ER & 100 & 50 & 0.28  & 0.24 & 0.39    & 0.36 & 0.44 & \textbf{0.46} \\ 
 ER & 200 & 100 & 0.29 & 0.35 & 0.62   & 0.60 & \textbf{0.63} & \textbf{0.63} \\ 
 
 Hub & 100 & 50 & 0.18 & 0.16 & 0.02     & 0.25 & \textbf{0.28} & 0.18 \\ 
 Hub & 200 & 100 & 0.15 & 0.19 & 0.01    & \textbf{0.24} & \textbf{0.24} & 0.10 \\ 
 \hline
\end{tabular}
\end{center}
\caption{\label{HighDim.tab}F1-score of conditional independence graph recovery using different model selection methods.  Results are the average over 50 replications when $N < p$.}
\end{table}

\section{Anomaly Detection} \label{sec:Anomaly}
Another application of our graph-coding approach is anomaly detection in multivariate Gaussian data. For detecting anomalous data, we utilize the atypicality concept developed in \cite{HostSabetiWalton15}. A sequence is atypical if it can be coded in fewer bits using a coder that is not the optimum coder for any typical sequence.
Theoretical properties and experimental practicality of atypicality were described in \cite{HostSabetiWalton15,Host16BD, host2018coding}. In the following, we describe the approach for using atypicality to find anomalous data in graph-based data.

\subsection{\label{AnomolousGraph.sec}Anomalous Graph Detection Algorithm}
In anomalous graph detection, we are given a set of $N$ i.i.d observations $\{\mathbf{x}_1, \ldots, \mathbf{x}_N\}$ from a $p-$variate Gaussian distribution as training set. The question is to determine whether a test set of $M$ i.i.d observations $\{\mathbf{y}_1, \ldots, \mathbf{y}_M\}$ generated from a different distribution, where $N \gg M$, is anomalous or not. 

The general procedure is as follows:
\begin{enumerate}
    \item On the set of training data $\{\mathbf{x}_1, \ldots, \mathbf{x}_N\}$, we use Algorithm~\ref{Alg} to find the best conditional independence graph $G_T$. It means that we learn the parameters of graph structure (e.g. degree distribution) that describes the data best. The typical coder is the estimated conditional independence graph structure $G_T$. Both encoder and decoder know the values of the parameters associated with $G_T$ (i.e. degree distribution) and thus, they do not need to be encoded.
    
    \item On the test data $\{\mathbf{y}_1, \ldots, \mathbf{y}_M\}$, we first compute the codelength of typical coder $L(G_T) + L(D|G_T)$ by computing the codelength of data $D$ and graph structure $G_T$ obtained from the previous step. We then use Algorithm~\ref{Alg} to find the best conditional independence graph $G_A$ describing the test data; here for each realization of $\lambda$. We also need to add the overhead of encoding the parameters associated with that specific conditional independence graph. The atypical codelength, $L(G_A) + L(D|G_A)$, is now the minimum of these codelengths. 
    
    \item The atypicality measure is the difference between the atypical codelength and the typical codelength. If $ L(G_A) + L(D|G_A) -  L(G_T) - L(D|G_T) < 0$, or when it is smaller than some threshold, then the test data is considered as anomalous.
\end{enumerate}
    
\subsection{Experiments}
We tested our method for anomaly detection on a real-world dataset. This dataset contains 12-lead ECG signals of a group of youth  16-21 years of age ($n=2366$). The dataset of 102 ECG variables was initially processed to select features with empirical distribution close to normal distribution. After screening all features and selecting those with approximately normal distribution, we ended up with 43 ECG variables for analysis.

The idea is that different age groups may show different patterns in their feature space, which may be too subtle to recognize by medical professionals or conventional ECG reading algorithms. Therefore, if we learn the underlying conditional independence graph for a specific age group, it might not perform well when it is applied to another age group. 
In the experiments, females of 16 years of age are considered as the typical data ($n=700$), and females of any other age group are considered as the atypical data ($n=1700$). It is expected that the universal source coder gives a better model for the atypical data, shorter codelength, when it is compared to the typical coder.

\begin{figure}[tbh]
\begin{centering}
\includegraphics[width= 3.1 in]{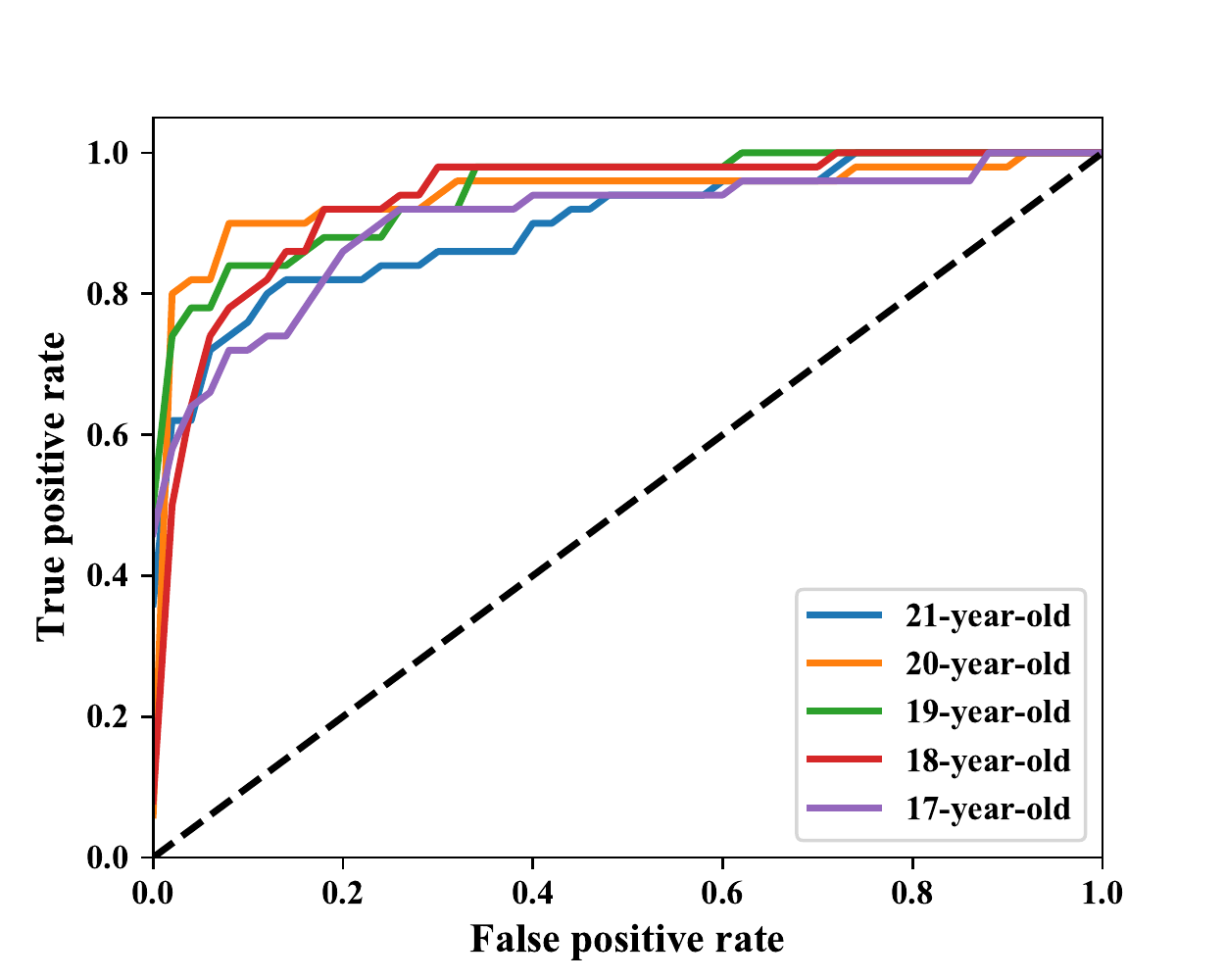}
\caption{\label{ECG_ROC.fig} ROC curve of atypicality measure when degree distribution is used. Each curve represents the measure when the corresponding age group is evaluated versus the 16-year-old group.}
\end{centering}
\end{figure}

To measure the performance, the receiver operating characteristic (ROC) curve and area under the curve (AUC) were used as the metrics. We estimated the typical coder based on 80\% of data on the typical data. Test data comprises of two different sets 1) 20\% of hold-out data from the typical dataset, 2) randomly chosen samples of the same size from each age group in the atypical data (one set for each age group). To compute the ROC curve, we repeated the experiment 50 times.
Figure~\ref{ECG_ROC.fig} depicts the ROC curve of the females of each age group versus 16-year-old females when we used degree distribution for coding the graph structure. Our methodology could distinguish a single age group (e.g. 17, etc. years) as atypical data compared to the training age-group (16 years) with a high level of accuracy. Despite the small age gap between the typical and atypical data and the fact that both datasets were from healthy individuals, there was a distinctive pattern of ECG variables that made the groups distinguishable by our novel application.  In comparison, current established ECG variables used to predict some heart condition by ECG provide an AUC of only 0.49-0.73 \cite{BratinscakWillamsAl15}.

One may ask how other graph coders perform on this task and we may not need to have more efficient graph coders. Our experiments show when we use IID structure coder, the performance drops compared to the case when we use degree distribution or triangle coder as given in Table~\ref{AD.tab}. It means that the coders based on degree distribution and triangle can extract more information from the graph and use it toward shorter codelength, which leads to better performance. 

\begin{table}
\begin{center}
\renewcommand{\arraystretch}{1.4}
\begin{tabular}{ |C{1.5cm}|C{1cm}|C{1cm}|C{1cm}| } 
 \hline
 Age Group &  Degree & IID & Triangle \\ 
 \hline
 17-year-old & \textbf{0.90} & 0.85 & 0.89 \\ 
 18-year-old & \textbf{0.92} & 0.90 & 0.90 \\ 
 19-year-old & 0.94 & 0.93 & \textbf{0.95} \\ 
 20-year-old & \textbf{0.93} & 0.92 & 0.92 \\ 
 21-year-old & \textbf{0.89} & 0.85 & \textbf{0.89} \\ 
 \hline
\end{tabular}
\end{center}
\caption{\label{AD.tab}AUC of atypicality measure for different graph coders.}
\end{table}

\section{Conclusion}
In this paper, we extended the description length analysis to graph-based data by encoding graph structure and data according to the graph structure. We introduced two applications: 1) model selection, 2) anomaly detection in Gaussian graphical models.
First, we demonstrated that our approach for model selection gives a more accurate graph model of data compared to commonly used methods that overlook the graph structure as an informative part in decision-making. While in most cases coding with degree distribution and triangle outperform coding with IID structure by a wide margin, there are few cases where IID structure offers a slight improvement. In \cite{Mojtaba21GraphCoding}, we will address this by combining degree distribution and triangle coders into a single coder that works in all cases.
Second, we utilized description length for anomaly detection by using codelength as the decision criteria. In the future, we will extend this approach to other types of graph-based data and applications.


\end{document}